\documentclass[runningheads]{llncs}
\usepackage{graphicx}
\begin{document}
\title{Evaluating the Explainability of Attributes and Prototypes for a Medical Classification Model}
\titlerunning{Evaluating the Explainability of Attributes and Prototypes}
%
\author{Luisa Gallée\inst{1}\orcidID{0000-0001-5556-7395} \and Catharina Silvia Lisson \inst{2} \and  Christoph Gerhard Lisson \inst{2} \and Daniela Drees\inst{2}\orcidID{0000-0002-3305-2945}  \and Felix Weig\inst{2} \and Daniel Vogele\inst{2}\orcidID{0000-0002-6421-1400} \and Meinrad Beer \inst{2,3}\orcidID{0000-0001-7523-1979} \and
Michael Götz\inst{1,3}\orcidID{0000-0003-0984-224X}}
\authorrunning{L. Gallée et al.}
%
\institute{Experimental Radiology, University Hospital Ulm, Germany \email{\{luisa.gallee,michael.goetz\}@uni-ulm.de} \and  Department of Diagnostic and Interventional Radiology, University Hospital Ulm, Germany  \and i2SouI - Innovative Imaging in Surgical Oncology Ulm, University Hospital Ulm, Germany}
\maketitle              
\begin{abstract}

Due to the sensitive nature of medicine, it is particularly important and highly demanded that AI methods are explainable.
This need has been recognised and there is great research interest in xAI solutions with medical applications.
However, there is a lack of user-centred evaluation regarding the actual impact of the explanations.
We evaluate attribute- and prototype-based explanations with the Proto-Caps model. This xAI model reasons the target classification with human-defined visual features of the target object in the form of scores and attribute-specific prototypes. The model thus provides a multimodal explanation that is intuitively understandable to humans thanks to predefined attributes.
A user study involving six radiologists shows that the explanations are subjectivly perceived as helpful, as they reflect their decision-making process. The results of the model are considered a second opinion that radiologists can discuss using the model's explanations. However, it was shown that the inclusion and increased magnitude of model explanations objectively can increase confidence in the model's predictions when the model is incorrect. 
We can conclude that attribute scores and visual prototypes enhance confidence in the model. However, additional development and repeated user studies are needed to tailor the explanation to the respective use case.

\end{abstract}

\section{Introduction}
There is great interest in the use of AI in the medical field. Medical applications include robotic surgery \cite{hashimoto2018artificial}, clinical decision support \cite{he2019practical}, and, especially with advances in computer vision, medical imaging and diagnostics \cite{he2019practical,esteva2021deep}. Expectations include increased safety and time efficiency in medical decision making \cite{amann2023expectations}. The high-stakes environment of medicine, where critical decisions are made, poses exceptional challenges for AI developers. If an incorrect AI decision goes unrecognized, the consequences can be dramatic. High prediction accuracy in extensive testing is only one way of assessing an AI model. In addition, the use of explainable AI can make an important contribution to the understanding and consequently the acceptance of computer-aided decision support systems. 

The choice of the explanatory method must be in accordance with the task and the human decision-making process. For the model to be understood and trusted, it is crucial that its explanation aligns with the user's intuition \cite{chen2022machineexplanations}. Medical personnel are trained to diagnose according to defined criteria. Therefore, it is intuitive that an assisting AI should also learn these criteria and base its prediction on them. In this way, the AI's information process mirrors the human process.

This work focuses on diagnosing medical images, specifically determining the malignancy of pulmonary nodules in CT images. Optical differentiation features used for diagnosis include contour and texture properties. We utilize this reasoning process and employ the AI model Proto-Caps \cite{gallee2023interpretable}. The model explains its output prediction by describing the visual appearance of the target object. The features that are decisive for radiologists were mapped to the describing reasoning of the model so that the model closely mimics the radiologists' decision-making process. As the attributes are defined by humans, this supervised approach skips the need for additional uncertainty in the interpretation of the explanation. In addition to a score-based attribute description, the model expands the explanation modality with visual exemplar prototypes that are attribute-specific, which is generally seen as an increase of explainability for complex decision processes \cite{reyes2020interpretability}. The model shows a high performance in target classification as well as in recognizing the appearance features \cite{gallee2023interpretable}. However, the evaluation of the level of explainability of this approach is technically limited and an holistic evaluation of the effectiveness of the explanation methods is only possible by including the persons receiving the explanation \cite{liao2021Humancentered,dominguez2019effect}. 

To the best of our knowledge, there have been no studies involving users that investigate the level of explainability of attribute- and prototype-based deep neural networks on images. Our work aims to contribute to this field of research by presenting a user study with a medical application. We applied Proto-Caps on a dataset to classify pulmonary nodules and presented the model output with the intrinsic reasoning to radiologic experts. Using a questionnaire, we investigated the influence of the explanations on the confidence of the radiologists in the model and their performance. The study provides insights for the development of attribute- and prototype-based explainable methods and their presentation to the user.

\section{Related Work}
Despite the growing number of explainable methods, the evaluation involving end-users of specific methods in the application is clearly sparse \cite{nauta2023anecdotal,rong2023towards}. Especially in the application of AI in medical image processing, there is a lack of comprehensive user surveys that go beyond single expert recommendations. This can be argued with the relatively high cost and effort of expert domain knowledge required. In the following, we would like to discuss works and their findings that provide a human-centered evaluation of explainable methods across application field and task.

Chen et al. \cite{chen2023understanding} examined the impact of feature and example-based explanations on participants' decision outcomes in a classification task of tabular and textual data. They concluded that prototypes "were less disruptive of people’s natural intuition about outcomes, they better promoted inductive reasoning about features and the decision task generally, and, in particular, they provided strong and accurate signals of prediction unreliability" \cite{chen2023understanding} compared to feature-based explanations.

Tabular data was also the focus of the study by Dieber et al. \cite{DIEBER2022143}. Here, the model-agnostic explainable method LIME \cite{ribeiro2016should} was applied and evaluated with users. The study showed that the output of the LIME algorithm was difficult to understand intuitively, which leads to the conclusion that detailed documentation and clarification of the output is necessary for a positive effect on the user experience and the model's level of explainability.

A whole collection of explainable methods were compared in a work by Silva et al. \cite{silva2023explainable}, in which the task also involved answering  commonsense reasoning question. A virtual agent gave its prediction using the following explainable modalities: template language, counterfactual, decision tree, probability scores, crowd-sourced, case-based, feature importance, or without any explanations. The user survey shows a correlation between the accuracy of the question-answer response and the participant trust with the choice of explanation modality. In particular, the model probability scores were rated as significantly less explainable than the other modalities, including case-related explainability.

Bansal et al. \cite{bansal2021does} also conducted a user study on the effect of explanations on user performance. The tasks included a text classification with highlighted key words as explanation and a question-and-answer task with manually generated justifications. In the study, explanations improved the user performance when the system was correct, but had a negative effect on accuracy on examples when the system was incorrect.

Papenmeier et al. \cite{papenmeier2019model} compares the influence of model accuracy performance and explanation fidelity on user trust in the model. In the task of identifying offensive texts, a higher model accuracy led to more trust among the participants, whereas the explainability by highlighting decisive words did not lead to an increase in trust. Furthermore, the addition of nonsensical explanations was found to be even potentially damaging to trust.

\section{Methods and Materials}
This paper examines the explainability of the xAI model Proto-Caps \cite{gallee2023interpretable}, which uses visual features of the target object to reason the classification of it. The visual features are represented in form of predefined appearance characteristics and are represented with exemplary visual images and their attribute scores. As the selection of the descriptive characteristics is predetermined by humans, this form of explanation is explicitly human understandable.

\subsection{Dataset}
For this study, the model was trained with the medical dataset Lung Image Database Consortium and Image Database Resource Initiative (LIDC-IDRI) \cite{armato_iii_lung_data_2015}. It contains thorax computed tomography (CT) scans and characterization of pulmonary nodules. Each nodule was assessed in regards of its malignancy and its visual appearance. The malignancy label is used as the target of the model classification. Descriptive visual attributes are used for the reasoning of the classification prediction. The attributes have been identified by radiologists as crucial for assessing malignancy and include sphericity (\textit{linear} or \textit{round}), margin (\textit{poorly defined} or \textit{sharp}), spiculation (\textit{none} or \textit{marked}), and texture (\textit{non-solid} or \textit{solid}), among others. The pre-processing steps were identical to those described in the Proto-Caps methodological paper \cite{gallee2023interpretable}, including cutting out the lung nodules from the CT images and averaging the annotated scores across the raters.

\subsection{xAI-Model}
Proto-Caps \cite{gallee2023interpretable} is a powerful model that is interpretable by design.  Through a hierarchical decision structure, the target classification is based on supervised-learned high-level features. The features are visualised using exemplary examples and justify the target prediction at the attribute level.

\begin{figure}[h]
\centering
\includegraphics[width=0.95\textwidth]{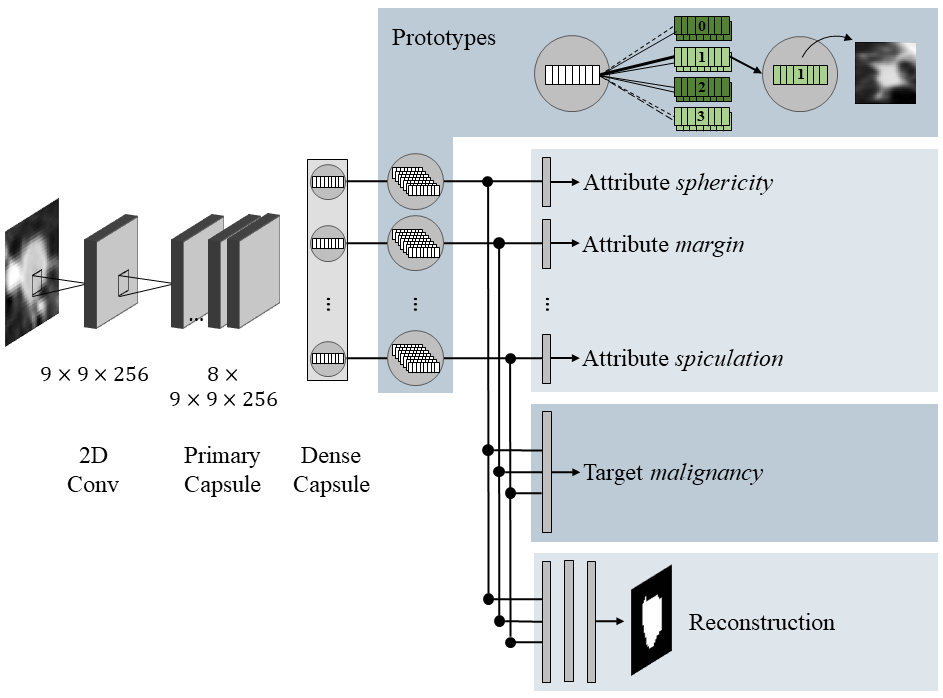}
\caption{\textbf{Proto-Caps architecture} A capsule network produces encapsulated representations of visually descriptive features. The prototype layer iteratively constructs a set of prototypes, each covering a single attribute. During inference, the latent vectors of the closest attribute prototypes are concatenated for a dense layer to predict a target score. The concatenated vector is also fed into a decoder network to reconstruct the region of interest, which benefits the training \cite{gallee2023interpretable}.} \label{fig:model}
\end{figure}

The model is based on a capsule network with a CNN backbone. Each capsule is mapped to a predefined attribute, and a prototype layer allows for the storage of 
typical examples for each capsule. Target classification is performed by concatenating the latent vectors of the closest prototypes per capsule. This inference procedure allows the prediction of the target to be reasoned proximately by the attribute predictions, which also reflects the way a human evaluates an object in an image. Figure \ref{fig:model} gives an overview of the hierarchical architecture of Proto-Caps. 
Details of the model architecture and training were adopted from the methodological paper \cite{gallee2023interpretable} and the code is publicly available at \url{https://github.com/XRad-Ulm/Proto-Caps}.
The model prediction accuracies were reported as $93\%$ for the target and $92\%$ on average for the attributes.

For the user study, cases were randomly selected from the test dataset, however ensuring a balance between cases in which the model was correct and incorrect. The output of the model comprises a prediction regarding malignancy and attribute scores, as well as visual attribute prototypes.

\section{Experiments}
As the model provides human-centered explanations, we conducted a survey with target users to assess the degree of explainability. Since expert knowledge is required to evaluate the medical data, we surveyed radiologists.

\subsection{Aim of study}
The experiment aims to evaluate the usefulness of the xAI method for end users. The following questions will be addressed.

\begin{quote}
    \begin{enumerate}
    \item[RQ1: ]How does the explanation affect the user's performance?
    \item[RQ2: ]How does the explanation affect the trust in the model?
    \item[RQ3: ]Are attribute-based explanation (scores, prototypes) helpful?
\end{enumerate}
\end{quote}

\subsection{Survey setup}
\begin{figure}[h]
\centering
\includegraphics[width=0.95\textwidth]{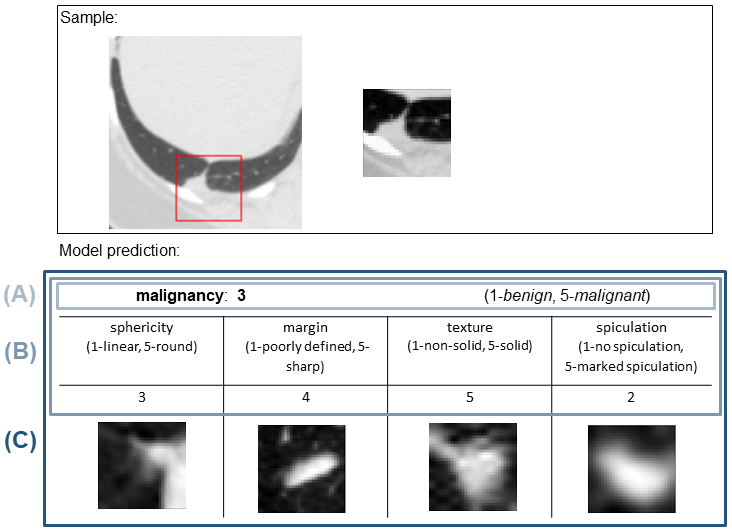}
\caption{\textbf{Test cases} Sample of test case in questionnaire. The pulmonary sample is shown in a section of the lung and in a crop-out view. Depending on the variant, \textbf{(A)} only the malignancy prediction of the model is given, \textbf{(B)} additionally the model predicted scores of the attribute, or \textbf{(C)} additionally prototypical samples of the attribute.} \label{fig:testcase}
\end{figure}

The experiment aims to investigate both objective and subjective perceptions of different modalities of descision explanation namely attribute scores and attribute prototypes. Open-ended and ordinal questions were used to assess radiologists' opinion toward the use of explainable AI in their field. In collaboration with an experienced radiologist, we developed a questionnaire organized into the following sections.
\begin{enumerate}
    \item Enquiring the participant's professional level as a radiologist
    \item Assessing the benefits and trust of AI and explainable AI in the radiological field (RQ2)
    \item Test cases: \\
    This section presents 36 test cases with sample lung nodules. In addition to the image of the sample nodule, the model output for the sample is also displayed to the participant. The model output is presented in three different variants:
    \begin{enumerate}
        \item[\textbf{(A)}] Malignancy prediction
        \item[\textbf{(B)}] Malignancy prediction + Attribute scores
        \item[\textbf{(C)}] Malignancy prediction + Attribute scores + Attribute prototypes
    \end{enumerate}
    Figure \ref{fig:testcase} displays the presentation of the sample nodule and model output.
    
    The samples were evenly divided, with 12 samples in each of the A, B, and C variants. Additionally, the model correctness was balanced in each category, resulting in nine samples being correctly predicted as benign, incorrectly predicted as benign, correctly predicted as malignant and incorrectly predicted as malignant.
    
    In addition to assessing malignancy (RQ1), participants can indicate their level of certainty and provide their confidence in the model's output (RQ2).
    \item Overall assessment of the helpfulness of the model explanations (RQ2, RQ3)    
\end{enumerate}

The questionnaires were distributed to the participants and could be completed without any time constraints. All participants consented to the analysis and publication of their questionnaire data.
    
\subsection{Participants}
The study participants were recruited from the University Hospital of Ulm, and six radiologists were enlisted to participate in the study. Four of the participants were senior radiologists with over 9 years of experience, while the remaining two were assistant physicians with 3 years of experience or less. Diagnosing CT images and classifying pulmonary nodules for malignancy was a daily task for five out of six participants and a weekly task for the sixth.
       
\section{Results and Discussion}

\subsection{Radiologists' attitude towards AI}
\subsubsection{What is the relevance of AI in radiologic image diagnosis, specifically in determining the malignancy of pulmonary nodules?}

In the presented task, the importance of AI assistance was rated as \textit{rather relevant} by four out of six participants, while two out of six participants rated it as \textit{highly relevant}.

The radiologists specified several reasons for their expectations. One aspect that was repeatedly mentioned was the increased safety of radiological diagnosis. It is expected that the assistance of AI support will increase sensitivity and reduce the likelihood of missed findings and incorrect decisions, which could have a significant impact on the patient's outcome.	

Detecting, segmenting and assessing nodules is a very time-consuming, repetitive task, and the participants hope to save time with AI-assisted radiological diagnosis. One participant even stated that the expected increase in workload can only be solved with the help of detection systems. As time savings go hand in hand with cost reductions, AI is of great interest in medicine \cite{khanna2022economics}.

\subsubsection{When is AI trustworthy?}\label{WhenisAItrustworthy}
When asked about criteria for the trustworthiness of AI models, two aspects were clearly emphasised by the radiologists: extensive performance testing and justification of the model's predictions.

Five of the six participants mentioned performance requirements, such as high sensitivity and specificity, and test conditions. One concern was the size and diversity of the test data set, such as the inclusion of different scanners from different vendors and test data from other institutions.

The importance of explanations for trustworthiness was emphasised by four of the six study participants. Reasoning and traceability of the AI's judgement in characterisation was considered particularly important. Comparative studies, such as radiologist performance with and without AI support, were mentioned as a possible evaluation.

\subsection{RQ1: How does the explanation affect the user's performance?}
\begin{figure}[h]
\centering
\includegraphics[width=1.0\textwidth]{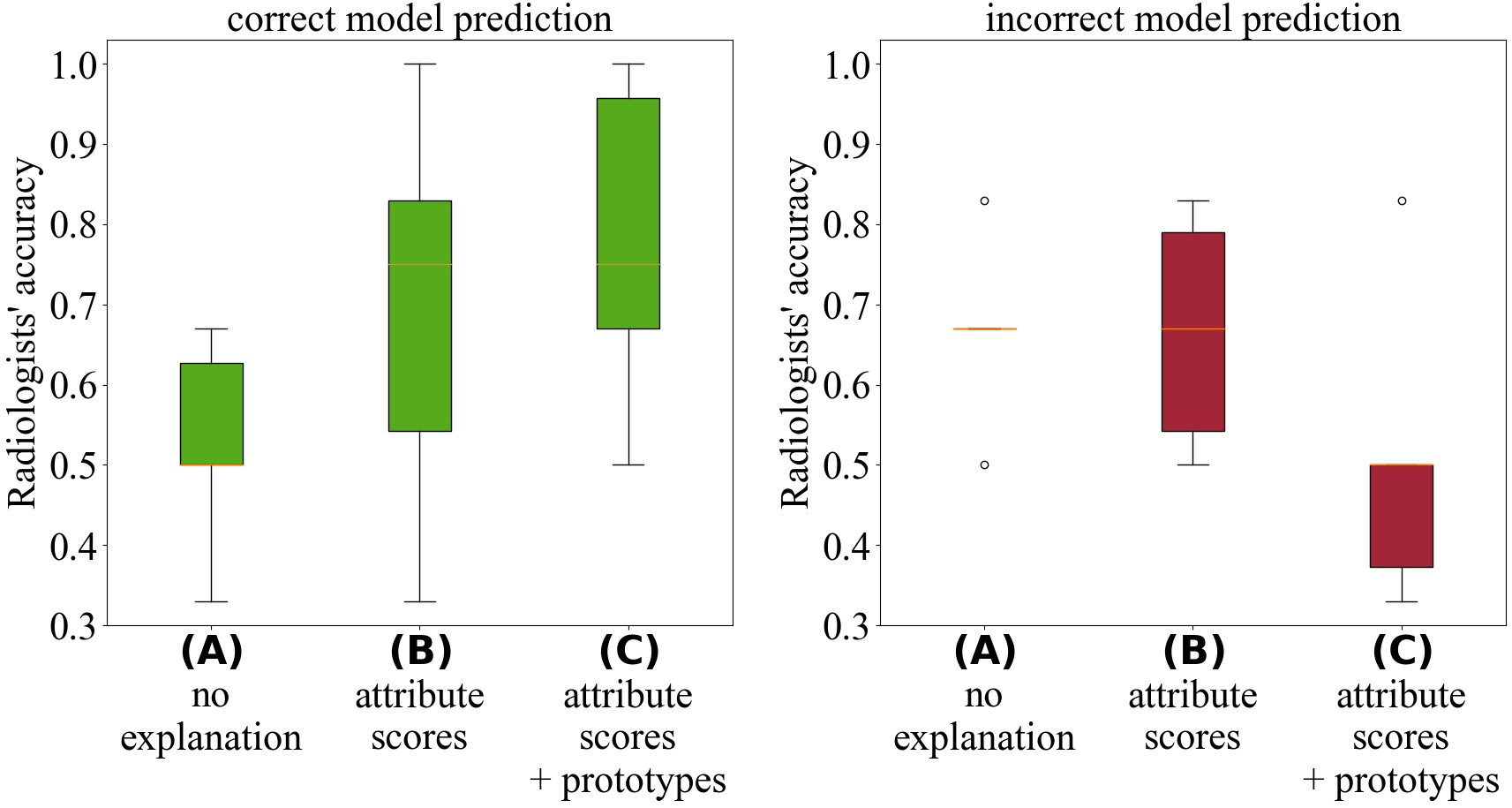}
\caption{\textbf{Users' performance} Analysis of the radiologists' performance during the test cases in Within-1-Accuracy \cite{gallee2023interpretable}. The boxplots in green depict the diagnostic accuracy of the radiologists when the model prediction was correct, while the red boxplots show the accuracy when the model was incorrect.} \label{fig:res:performance}
\end{figure}

In the test cases, the radiologists diagnosed the malignancy of the lung nodules presented. Figure \ref{fig:res:performance} displays the radiologists' performance according to the model output variant (\textbf{(A)} no explanation, \textbf{(B)} only attribute scores, \textbf{(C)} plus attribute prototypes) and the correctness of the model output. The ratio between test cases with correct and incorrect model prediction is balanced and does not reflect the actual performance of the model.

If the model prediction is correct, radiologists' performance improves as the model becomes more explainable. Radiologists perform best when the model explains predictions using attribute scores and prototypes.

On the other hand side, when the model output is incorrect, the user's performance is lower, when the model reasons its prediction with attribute scores and prototypes. If only the attribute scores are provided as an explanation, the radiologists' performance is similar to when no explanation is given.
\\
\\
In our study, the explanations had a marked influence on the radiologists' decision. While they can have a positive effect on user accuracy if the model is correct, they can also have a negative effect by convincing the user of an incorrect prediction. This finding aligns with existing literature on other explainable methods \cite{bansal2021does}, where a user study found that when using model explanations the performance improved in cases where the model was correct, but decreased it when the model prediction was incorrect.

However, it is important to acknowledge that a limited presentation of the test cases in the questionnaire may have affected the diagnostic accuracy of the radiologists, which could also affect their response to the AI. It is worth noting that, unlike in reality, the radiologists only had access to a limited information base for their diagnosis. Where they would otherwise have additional background information about the patient and the course of the disease for the diagnosis of lung nodules, they carried out the diagnosis in the test cases solely on the basis of a CT image, which leads to additional uncertainty. Additionaly, the radiologists stated that the visualization of the sample nodules was unusual for them. They specialise in evaluating lung nodules in 3D and with windowing functions in a PACS system. However, in the questionnaire they only had a two-dimensional slice image at their disposal.
The reduced self-confidence may have led to a greater willingness to be influenced by a reasoned model prediction.

\subsection{RQ2: How does the explanation affect the trust in the model?}

\begin{figure}[h]
\centering
\includegraphics[width=0.8\textwidth]{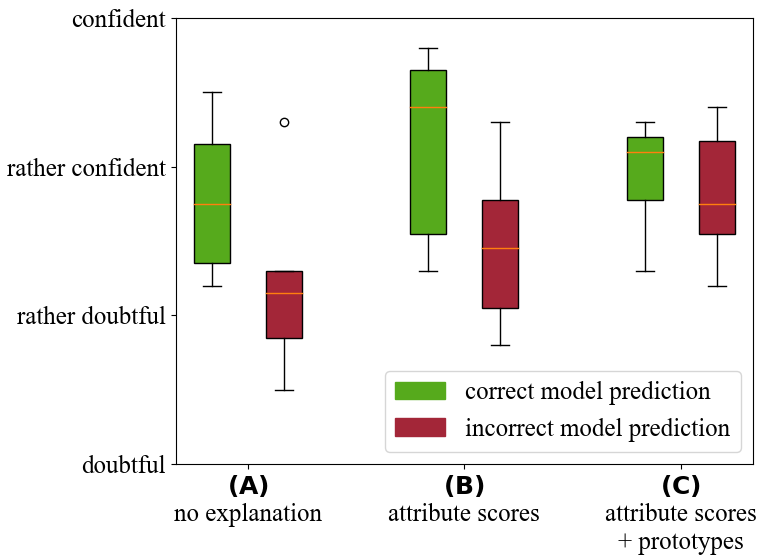}
\caption{\textbf{Trust in model prediction} The confidence scores in the model predictions during test cases are presented with respect to the model's correctness and level of explainability.} \label{fig:res:confidence}
\end{figure}

Figure \ref{fig:res:confidence} shows the confidence ratings of the participants in the assessment of the test cases. The participants are on average \textit{rather confident} in a correct model output, regardless of whether and to what extent the model explains its prediction. However, when analyzing samples with incorrect model predictions, there is a tendency for confidence in the model to increase with the model's explanation.
\\
\\
Our user study indicates that trust in the model is influenced by both the accuracy of the model and the extent of its reasoning.
When the model is wrong, the more explanations the model provides, the higher the confidence in the model.
The connection between the latter has also been established in earlier literature \cite{papenmeier2019model}. The confidence measure indicated a higher score in the cases with explanations than without. Additionally, participants demonstrated a greater willingness to accept the model's incorrect predictions. This aligns with the psychological literature \cite{koehler1991explanation}, which suggests that providing an explanation for a hypothesis increases people's confidence in its validity.

\subsection{RQ3: Are attribute-based explanation (scores, prototypes) helpful?}
\subsubsection{Do explainable methods facilitate radiologists' work?}
After completing the test cases, in which the participants were given an impression of working with the model as an assistant, the helpfulness of the explanation methods was rated. When participants were asked if the \textbf{explanations helped in determining malignancy}, five out of six participants for the attribute scores and three out of six participants for the attribute prototypes answered \textit{rather yes}. In addition, three out of six participants in the attribute scores and four out of six participants in the attribute prototypes answered the question of whether the \textbf{explanations facilitated their work} with \textit{rather yes}. The radiologists see advantages in the model explanations and support their ratings with the following arguments.

Both variants of the model explanations are considered a second opinion, either to confirm one's own assessment or to prompt a reassessment. The positive emphasis was placed on numerical attribute explanations, as they provide support for intuition. The prototypes were also positively received as they aid in understanding the AI model's thought process.

A second argument that has been mentioned repeatedly is a guidance effect on the differentiation criteria. Since the attributes shown are consistent with those that are also crucial for radiologists, both explanation variants were seen as helpful reminders of the malignancy criteria. For the prototypes, it was noted that the "examples can help to focus the radiologist's eye once again on certain areas or parts of the lung findings that were not taken into account when the image was first viewed". 

Four out of six participants have expressed a need for additional information for real-life diagnosis of lung nodules, which should also be included in the model. The factors considered for nodule differentiation also include size, clinical parameters, tumour and surgery history, disease progression, and risk factors such as age and smoking history. In particular, it was pointed out that the model does not currently include the position of the nodule within the lung, which is also a differentiation criterion for radiologists. As a further optimisation of the model explanation, the prediction certainty was mentioned. This is consistent with research that shows a combination of explanatory modalities improves the validation basis of the model's predictions \cite{reyes2020interpretability}.

\subsubsection{Are explainable methods beneficial in training radiology beginners?}
Three out of six radiologists agree with the statement that beginner radiologists would benefit from model explanations. The differentiation criteria taught in radiology training include those of the model and are also demonstrated with exemplary images. It has been noted that the AI model's visualization of the signs of malignancy can be helpful for inexperienced radiologists.

However, the interviewees recognise the potential danger of relying too heavily on AI-supported diagnostics, particularly for inexperienced radiologists. It was noted that the model only receives the image as input, which is insufficient for a reliable diagnosis compared to a comprehensive patient view that includes all relevant information. When assessing the model, the AI user must be aware of this.

\subsection{Limitations}
The limitations of the questionnaire arise from its being created outside the specific radiological diagnostic software. As a result, the presentation of the test cases was restricted. The sample nodules were displayed in two dimensions and did not correspond to the three-dimensional CT image that radiologists are accustomed to. Furthermore, radiology display programs provide the option to adjust the Hounsfield unit (HU) display, which was set to a fix lung window in the questionnaire. The display limitations also applies to the example nodules for the attribute prototypes.

Furthermore, radiologists may find it difficult to evaluate lung nodules without additional information on progression and the patient. Because of these limiting factors, the diagnostic accuracy of radiologists cannot be realistically assessed.

In addition to the limitations in presenting the questionnaire, it is important to mention the limited number of study participants. Conducting studies with end-users in the medical field is expensive due to the high level of expertise required. With only six radiologists, this study can only serve as a primary investigation into the use of explainable AI. Developing an application-specific tailored solution requires multiple iterations with user studies and a larger testing group. 

\section{Conclusion}
Motivated by the lack of human-centred evaluation of explainable image processing methods, in this paper we evaluated the helpfulness of providing visual characteristics as explanations for a medical image classification task. In particular, we investigated the influence of model explanation on trust in the model. The explainable method investigated focussed on humam-defined attributes and was represented by the inherently explainable model Proto-Caps through scores and prototypical examples. To assess the usefulness of the model explanations, six radiology experts were surveyed using a questionnaire and application tasks.

The survey showed that all radiologists consider the use of AI in the diagnosis of radiological images to be relevant and forward-looking. In addition to high performance in a carefully designed test environment, the majority of radiologists consider the explanation of the model output to be an important criterion for trustworthiness.

Subjectively, the explanations presented are considered helpful because they cover the same features as the differentiation criteria used by radiologists. The model results are seen as a second opinion that they discuss based on the multimodal representation of the explanations. 

There is an objective tendency to over-rely on the model when it provides explanations, which is consistent with findings from psychology. If the model prediction is incorrect, the model is more trusted if it explains its prediction than if it gives no explanation. 
 
For future work it is worth investigating in the presentation of the model output, including its explanations, to discourage overconfidence in the model prediction and decreasing the psychological tendency of falling into the misleading fallacy that a model's explanatory power equals its accuracy.

This work demonstrates the importance of involving the target users of AI methods in their development. The need for explainable AI methods in medical applications is high, and the evaluation of these methods is important to assess and improve their helpfulness.

\section*{Acknowledgment}
We thank all participants of the University Hospital Ulm for their contribution to the user study.


\begin{thebibliography}{8}

\bibitem{hashimoto2018artificial}
Hashimoto, D. A., Rosman, G., Rus, D., Meireles, O. R.: Artificial intelligence in surgery: promises and perils. Annals of surgery vol. 268, nr. 1, pp. 70--76 (2018) \doi{10.1097/SLA.0000000000002693}

\bibitem{he2019practical}
He, J., Baxter, S. L., Xu, J., Xu, J., Zhou, X., Zhang, L.: The practical implementation of artificial intelligence technologies in medicine. Nature Medicine vol. 25, nr. 1, pp. 30-–36 (2019) \doi{10.1038/s41591-018-0307-0}

\bibitem{esteva2021deep}
Esteva, A., Chou, K., Yeung, S., Naik, N., Madani, A., Mottaghi, A., Liu, Y., Topol, E., Dean, J., Socher, R.: Deep learning-enabled medical computer vision vol. 4, nr. 1, pp. 5 (2021) \doi{10.1038/s41746-020-00376-2}

\bibitem{amann2023expectations}
Amann, J., Vayena, E., Ormond, K. E., Frey, D., Madai, V. I., Blasimme, A.:Expectations and attitudes towards medical artificial intelligence: A qualitative study in the field of stroke. Plos one vol. 18, nr. 1 (2023) \doi{10.1371/journal.pone.0279088}

\bibitem{chen2022machineexplanations}
Chen, C., Feng, S., Sharma, A., Tan, C.: Machine explanations and human under-standing (2022) \doi{10.48550/arXiv.2202.04092}

\bibitem{gallee2023interpretable} L. Gallée, M. Beer and M. Götz, ``Interpretable Medical Image Classification Using Prototype Learning and Privileged Information,'' in {\it Proc. MICCAI,}Vancouver, BC, Canada,  2023,
pp. 435--445, \doi{10.1007/978-3-031-43895-0\_41}

\bibitem{reyes2020interpretability}
Reyes, M., Meier, R., Pereira, S., Silva, C. A., Dahlweid, F. M., Tengg-Kobligk, H., Summers, R. M., Wiest, R.: On the interpretability of artificial intelligence in radiology: challenges and opportunities. Radiology: artificial intelligence vol. 2, nr. 3 (2020) \doi{10.1148/ryai.2020190043}

\bibitem{liao2021Humancentered} 
Liao, Q.V., Varshney, K.R.: Human-centered explainable AI (XAI): from algorithms to user experiences (2021) \doi{10.48550/arXiv.2110.10790}

\bibitem{dominguez2019effect}
Dominguez, V., Messina, P., Donoso-Guzman, I., Parra, D.: The effect of explanations and algorithmic accuracy on visual recommender systems of artistic images, in {\it Proc. ACM IUI} Los Angeles, CA, USA, 2019, pp. 408--416 \doi{10.1145/3301275.3302274}

\bibitem{nauta2023anecdotal}
Nauta, M., Trienes, J., Pathak, S., Nguyen, E., Peters, M., Schmitt, Y., Schlötterer, J., van Keulen, M., Seifert, C.: From anecdotal evidence to quantitative evaluation methods: A systematic review on evaluating explainable ai. ACM Computing Surveys vol. 55, nr. 13s, pp. 1--42 (2023) \doi{10.1145/3583558}

\bibitem{rong2023towards}
Rong, Y., Leemann, T., Nguyen, T., Fiedler, L., Qian, P., Unhelkar, V., Seidel, T., Kasneci, G., Kasneci, E.: Towards human-centered explainable ai: A survey of user studies for model explanations. IEEE Transactions on Pattern Analysis and Machine Intelligence 1--20 (2023) \doi{10.1109/TPAMI.2023.3331846}

\bibitem{chen2023understanding}
Chen, V., Liao, Q. V., Wortman Vaughan, J., Bansal, G.: Understanding the role of human intuition on reliance in human-AI decision-making with explanations, in {\it Proc. ACM CHI} Hamburg, Germany, 2023, vol. 7, nr. CSCW2, pp. 1--32 \doi{10.1145/3610219}

\bibitem{DIEBER2022143}
Dieber, J., Kirrane, S.: A novel model usability evaluation framework (MUsE) for explainable artificial intelligence. Information Fusion vol. 81, pp. 143--153 (2022) \doi{10.1016/j.inffus.2021.11.017}

\bibitem{ribeiro2016should}
Ribeiro, M. T., Singh, S., Guestrin, C.: "Why should I trust you?" Explaining the predictions of any classifier in {\it Proc. ACM SIGKDD}, San Francisco, CA, USA, 2016, pp. 1135--1144 \doi{10.1145/2939672.2939778}

\bibitem{silva2023explainable}
Silva, A., Schrum, M., Hedlund-Botti, E., Gopalan, N., Gombolay, M.: Explainable artificial intelligence: Evaluating the objective and subjective impacts of xai on human-agent interaction. International Journal of Human--Computer Interaction vol. 39, n. 7, pp. 1390--1404 (2023) \doi{10.1080/10447318.2022.2101698} 

\bibitem{bansal2021does}
Bansal, G., Wu, T., Zhou, J., Fok, R., Nushi, B., Kamar, E., Ribeiro, M. T., Weld, D.: Does the Whole Exceed its Parts? The Effect of AI Explanations on Complementary Team Performance in {\it Proc. ACM CHI} Yokohama, Japan, 2021, nr. 81,  pp. 1--16 \doi{10.1145/3411764.3445717}

\bibitem{papenmeier2019model}
Papenmeier A., Englebienne, G., Seifert, C.: How model accuracy and explanation fidelity influence user trust (2019) \doi{https://doi.org/10.48550/arXiv.1907.12652}

\bibitem{armato_iii_lung_data_2015}S. G. Armato III, G. McLennan, L. Bidaut, M. F. McNitt-Gray, C. R. Meyer, A. P. Reeves, B. Zhao, D. R. Aberle, C. I. Henschke, E. A. Hoffman and others , ``Data From LIDC-IDRI,''{\it TCIA,} (2015) \doi{10.7937/K9/TCIA.2015.LO9QL9SX}

\bibitem{khanna2022economics}
Khanna, N .N., Maindarkar, M. A., Viswanathan, V., Fernandes, J. F. E, Paul, S., Bhagawati, M., Ahluwalia, P., Ruzsa, Z., Sharma, A., Kolluri, R., and others: Economics of Artificial Intelligence in Healthcare: Diagnosis vs. Treatment. Healthcare, vol. 10, nr. 12 (2022) \doi{10.3390/healthcare10122493}

\bibitem{koehler1991explanation}
Koehler, D. J.: Explanation, imagination, and confidence in judgment. Psychological bulletin, vol. 110, nr. 3, pp. 499-–519 (1991) \doi{10.1037/0033-2909.110.3.499}

\end{thebibliography}
\end{document}